\documentclass[10pt,twocolumn,letterpaper]{article}

\usepackage[pagenumbers]{cvpr}      
\usepackage{multirow}

\definecolor{cvprblue}{rgb}{0.21,0.49,0.74}
\usepackage[pagebackref,breaklinks,colorlinks,allcolors=cvprblue]{hyperref}

\def\paperID{*****} 
\def\confName{CVPR}
\def\confYear{2025}

\title{GazeNLQ @ Ego4D Natural Language Queries Challenge 2025}

\author{Wei-Cheng Lin\textsuperscript{1}\thanks{equal contributions.}, Chih-Ming Lien\textsuperscript{1}\footnotemark[1], Chen Lo\textsuperscript{1}, Chia-Hung Yeh\textsuperscript{1, 2}\\ 
\textsuperscript{1}National Taiwan Normal University \hspace{2mm} \textsuperscript{2}National Sun Yat-sen University \\
{\tt\small \{linwc510, lien1119, chyeh, clo20\}@ntnu.edu.tw}
}

\begin{document}
\maketitle
\begin{abstract}
This report presents our solution to the Ego4D Natural Language Queries (NLQ) Challenge at CVPR 2025. Egocentric video captures the scene from the wearer's perspective, where gaze serves as a key non-verbal communication cue that reflects visual attention and offer insights into human intention and cognition. Motivated by this, we propose a novel approach, GazeNLQ, which leverages gaze to retrieve video segments that match given natural language queries. Specifically, we introduce a contrastive learning-based pretraining strategy for gaze estimation directly from video. The estimated gaze is used to augment video representations within proposed model, thereby enhancing localization accuracy. Experimental results show that GazeNLQ achieves R1@IoU0.3 and R1@IoU0.5 scores of 27.82 and 18.68, respectively. Our code is available at \url{https://github.com/stevenlin510/GazeNLQ}.
\end{abstract}    
\section{Introduction}
\label{sec:intro}
The goal of the Ego4D \cite{Ego4D} Natural Language Queries (NLQ) challenge is to temporally localize the segment of egocentric video that corresponds to a given natural language query. Existing approaches generally fall into two categories: pretraining a foundation model to learn transferable representations suitable for various downstream tasks \cite{InternVideo2022,EgoVLP2022, EgoVLPv2_2023, Egovideo2024}, or developing specialized grounding model tailored to the NLQ task\cite{Reler2022, groundnlq2023, objectnlq2024}.

Pretraining foundation models on large-scale dataset has yielded impressive results on numerous downstream tasks. For instance, InternVideo \cite{InternVideo2022} explores three types of feature extractors as backbone and fine-tunes them on the Ego4D training set. EgoVLP \cite{EgoVLP2022} constructs a large-scale egocentric training dataset and adapts video-text contrastive learning to explore representations. EgoVideo \cite{Egovideo2024} enhances training data quality by filtering and selecting samples from multiple existing datasets, leveraging video-text contrastive learning for model training. Alternatively, task-specific models such as GroundNLQ \cite{groundnlq2023} adopt a two-stage pretraining strategy framework and introduces a multi-modal multi-scale grounding module that enables early fusion of video and text features. ObjectNLQ \cite{objectnlq2024} enhances video representation by incorporating object-level information extracted through an object detection model.

Despite these advancements, most methods focus on visual and textual modalities, with limited exploration of auxiliary sensor data such as head motion or gaze signals in egocentric video understanding. Recently, EgoDistill \cite{egodistill} demonstrated the utility of head motion signals captured by the inertial measurement unit (IMU) of a head-mounted camera to facilitate efficient egocentric video understanding. Given that IMU data has been shown to improve classification accuracy in egocentric action recognition, it raises the question of whether the characteristics of egocentric video can similarly benefit egocentric video-language grounding. In egocentric video, gaze aligns closely with the camera wearer's field of view, serving as a natural and informative cue for providing valuable information about visual attention, cognitive process, and underlying intentions. Understanding gaze behavior is essential for many applications, including cognitive science and psychology, human-robot interaction, and virtual and augmented reality. Recognizing the central role of gaze in revealing attention and intention in egocentric contexts, we aim to leverage this cue to advance understanding in egocentric video analysis. Therefore, We propose GazeNLQ, a novel framework that incorporates gaze to enhance natural language grounding in egocentric videos. We use a contrastive learning strategy to train the gaze estimator, which predicts gaze directly from video. The estimated gaze is then used to augment video features, leading to promising results on the NLQ task.
\begin{figure*}[t]
    \centering
    \includegraphics[width=0.8\textwidth]{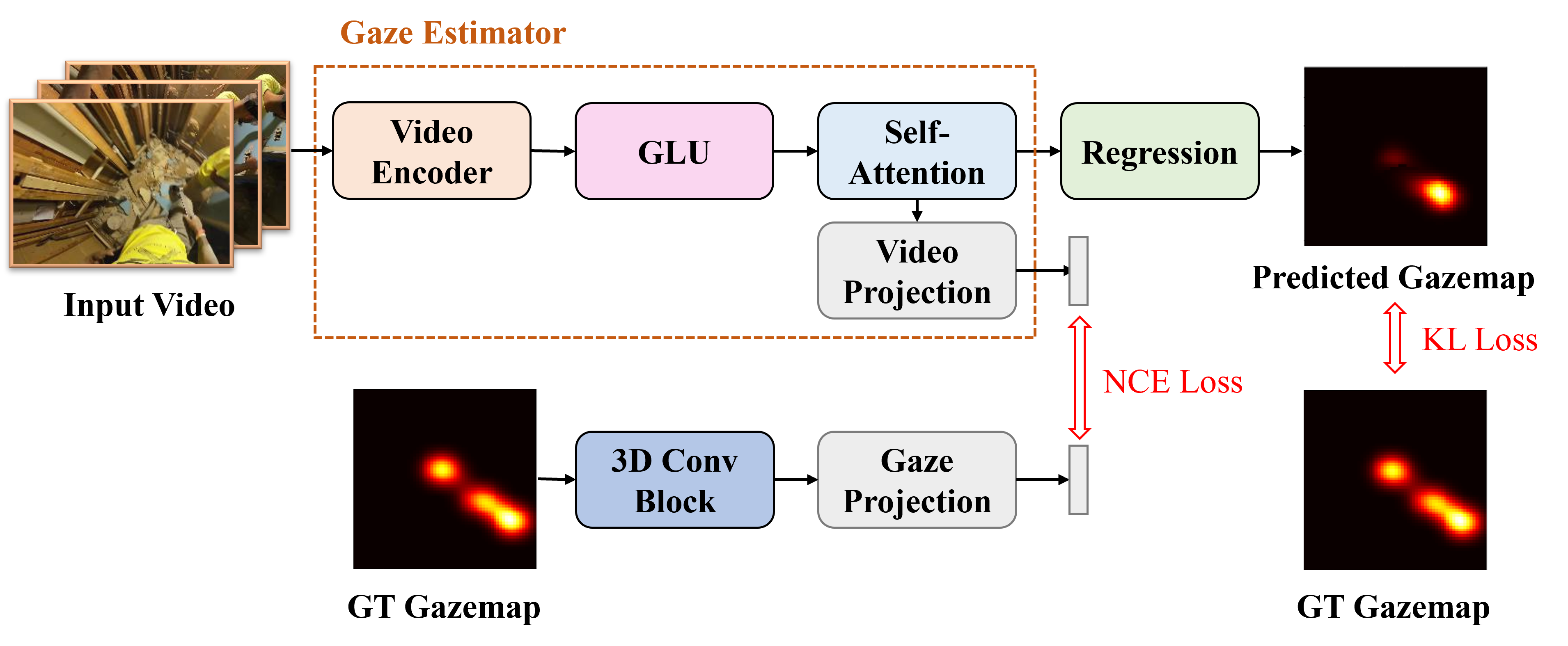}
    \caption{The proposed training framework for gaze estimator using contrastive learning.}
    \label{fig:gaze_estimator}
\end{figure*}

\begin{figure*}[t]
    \centering
    \includegraphics[width=\textwidth, keepaspectratio]{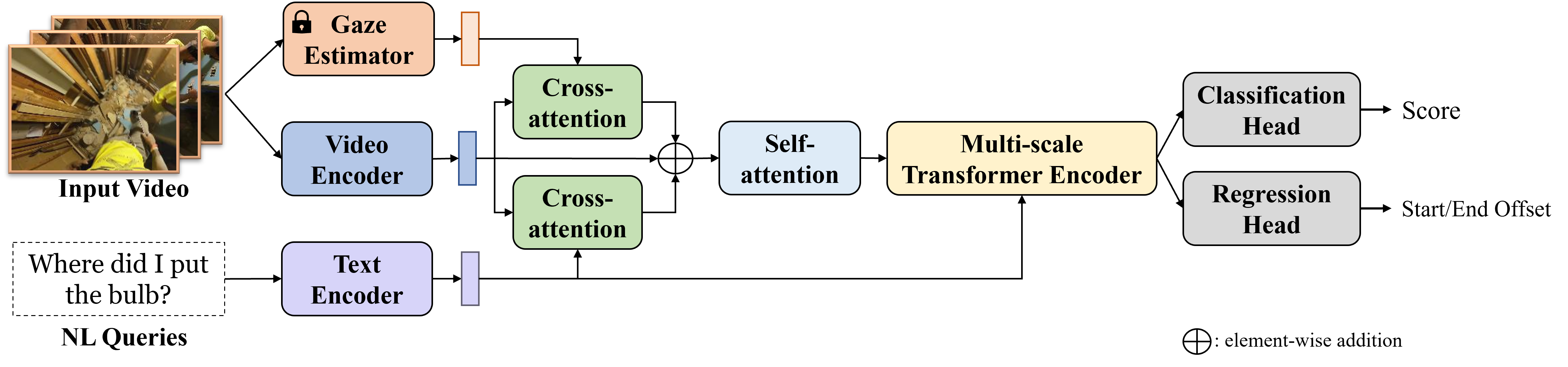}
    \caption{The proposed model for video temporal grounding.}
    \label{fig:architecture}
\end{figure*}
\section{Method}

This section presents GazeNLQ detailing the multi-modal feature representation and proposed model architecture. 

\subsection{Multi-modal Feature Representation} 
\textbf{Text and Video Representation.}
Following GroundNLQ \cite{groundnlq2023}, We extract textual token representations using the CLIP \cite{CLIP} text encoder and construct video representation by concatenating features from InterVideo \cite{InternVideo2022} and EgoVLP \cite{EgoVLP2022}. \\

\noindent\textbf{Gaze Representation.}
Since gaze annotations are not available for all video in the NLQ dataset, we train a gaze estimator using only the annotated data. The gaze estimator directly estimates the gaze from video. Our approach utilizes the dual-branch structure and contrastive learning for training, as illustrated in \cref{fig:gaze_estimator}. The architecture includes a video encoder, 5 Gated Linear Unit (GLU) layers, an self-attention layer, and a video projection head. The video features are first extracted from Omnivore \cite{omnivore} video encoder, then processed through the GLU and an attention layer before being projected into an aligned gaze embedding space. For the gaze branch, we follow the preprocessing procedure provided by \cite{Gaze} to generate the gaze map for each frame from the raw gaze data. These gaze maps are processed through a 3D convolution block and a gaze projection head to produce corresponding gaze embeddings. To align video embeddings with gaze embeddings, we employ the contrastive loss:
\begin{equation}
\mathcal{L}_{\text{NCE}}=\sum_i-\log\frac{\exp ({v_i\cdot g_+}/{\tau})}{\sum_j\exp(v_i \cdot g_j/{\tau})},
\end{equation}
where $v_i$ is video embedding, $g_+$ is the positive gaze embedding, $g_j$ is negative samples and $\tau$ is a temperature hyperparameter.
Additionally, a regression module predicts the gaze map $G_{\text{pred}}$, which is compared to the ground truth $G_{\text{GT}}$ using the KL divergence loss:
\begin{equation}
    \mathcal{L}_{\text{KL}} = D_\text{KL}(G_{\text{GT}} \parallel G_{\text{pred}}),
\end{equation}
where the $D_\text{KL}$ denote the KL divergence between $G_{\text{GT}}$ and $G_{\text{pred}}$. The total loss $\mathcal{L}_{\text{gaze}}$ is defined as:

\begin{equation}
    \mathcal{L}_{\text{gaze}}=\mathcal{L}_{\text{NCE}}+\mathcal{L}_{\text{KL}}.
\end{equation}

\subsection{Model Architecture}

The overall architecture of the proposed method is illustrated in \cref{fig:architecture}. The framework extracts gaze, video, and textual embeddings using a gaze estimator, a video encoder, and a text encoder, respectively. Two cross-attention modules are then employed to align and integrate the gaze and text embeddings with the video embeddings. The resulting embeddings are combined via element-wise addition and further refined using a self-attention. Next, we leverages the multi-scale transformer encoder architecture introduced in \cite{groundnlq2023} to enhance the modeling of hierarchical and temporal dependencies. Final predictions are produced by the classification head, which scores each interval in the feature pyramid, and a regression head, which estimate the boundary distances from the interval, similar to the approach described in \cite{groundnlq2023}. Model training employs the binary classification loss $\mathcal{L}_{\text{cls}}$ and Intersection over Union (IoU) regression loss $\mathcal{L}_{\text{reg}}$. The total loss of video temporal grounding $\mathcal{L}_{\text{localization}}$ is defined as:

\begin{equation}   \mathcal{L}_{\text{localization}}=\mathcal{L}_{\text{cls}}+\mathcal{L}_{\text{reg}}.
\end{equation}
\section{Experiment}

\begin{figure}
    \centering
    \includegraphics[width=1.0\linewidth]{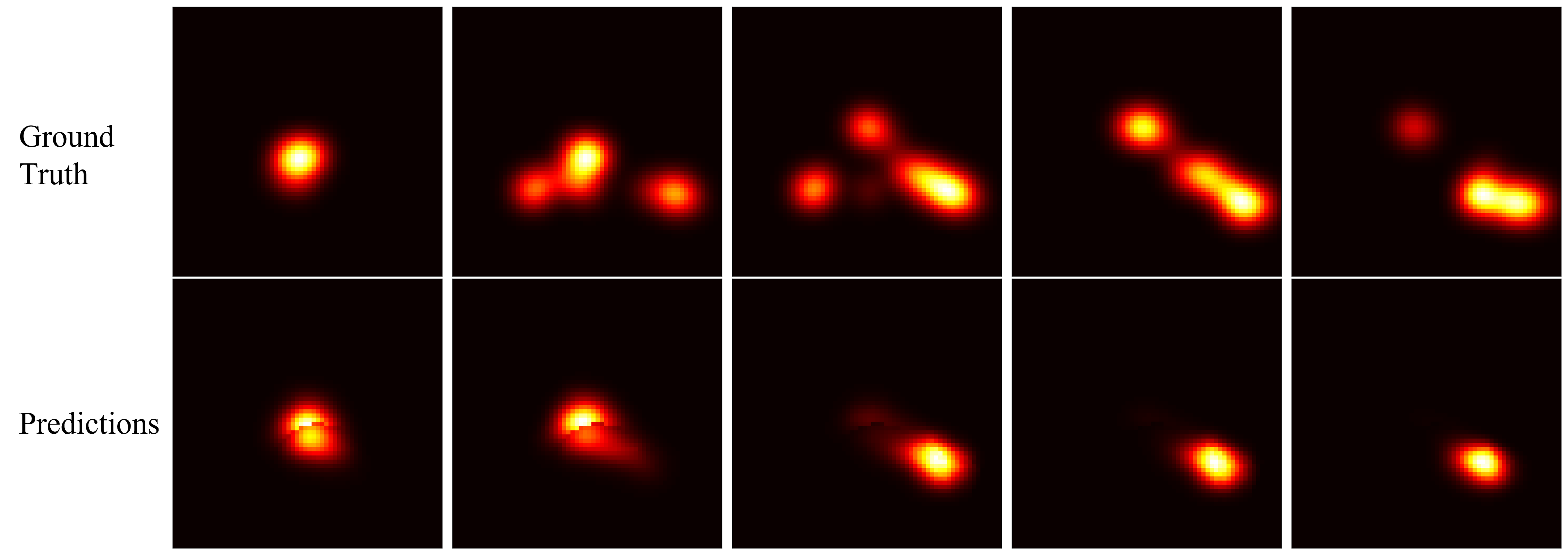}
    \caption{Visualization of gaze estimation. The top row shows the ground-truth gaze heatmaps, while the bottom row shows the predicted heatmaps.}
    \label{gaze_com}
\end{figure}

\subsection{Implementation Details}
\textbf{Gaze Estimator Training}. We begin by pretraining a gaze estimator using features extracted from the pretrained Omnivore model \cite{omnivore}. Omnivore processes video segments with a window size of 32 frames and a stride of 16 frames, , yielding a single feature vector per temporal window. To align the ground-truth supervision with this temporal resolution, we average the corresponding gaze heatmaps over each 32-frame segment at a resolution of $64 \times 64$, as illustrated in \cref{gaze_com}. For contrastive learning, both video and gaze representations are projected into embedding size of 384, which aligns with the dimensionality of the subsequent finetuning stage. The gaze estimator is trained using a learning rate of $1 \times 10^{-3}$ with a batch size of 16.\\

\noindent\textbf{Grounding Model Finetuning}. We adopt the GroundNLQ architecture \cite{groundnlq2023} and initialize it with pretrained weights from a model trained on narration data \cite{naq} to establish a strong starting point. Following pretraining of the gaze estimator, we incorporate it into the GroundNLQ pipeline for end-to-end finetuning. Additionally, we investigate a model variant called GazeNLQ\textsuperscript{$\star$} that employs negative gaze embedding, directing the model's attention to regions outside the gaze area. During this phase, we freeze the gaze estimator's weights and train the combined model for ten epochs, incorporating a warm-up period of four epochs. For the finetuning process, we utilize a learning rate of $2.5 \times 10^{-5}$ and a batch size of 8. All experiments are conducted using a single NVIDIA RTX 4090 GPU. During inference, we apply Soft-NMS \cite{softnms} to merge overlapping moment predictions, optimizing the final localization outputs. \\

\noindent\textbf{Ensemble}. We combines predictions from GroundVQA \cite{groundvqa}, which followed the strategy by EgoVideo \cite{Egovideo2024}. GroundVQA incorporates the question-answering data into the video grounding task by using the large language model. 

\begin{table}[t]
  \centering
  \fontsize{8}{9.5}\selectfont
  \caption{Performance comparison on NLQ \textit{test} split.}
  \begin{tabular}{@{}lc@{}lc@{}lc@{}lc@{}}
    \toprule
    \multicolumn{1}{c}{\multirow{2}{*}{Method}} & \multicolumn{4}{c}{Test Private}\\
    \cmidrule(lr){2-5} 
    \multicolumn{1}{c}{} & \multicolumn{1}{c}{R1@0.3} & \multicolumn{1}{c}{R1@0.5} & \multicolumn{1}{c}{R5@0.3} & \multicolumn{1}{c}{R5@0.5} \\
    \midrule   
    GroundNLQ \cite{groundnlq2023}  &  \multicolumn{1}{c}{24.50}   & \multicolumn{1}{c}{17.31}  & \multicolumn{1}{c}{40.46}  & \multicolumn{1}{c}{29.17} \\
    GroundNLQ$^{\dag}$ \cite{groundnlq2023}  &  \multicolumn{1}{c}{25.67}   & \multicolumn{1}{c}{18.18}  & \multicolumn{1}{c}{42.05}  & \multicolumn{1}{c}{29.80} \\
    ObjectNLQ$^{\dag}$ \cite{objectnlq2024}  &  \multicolumn{1}{c}{27.02}   & \multicolumn{1}{c}{19.28}  & \multicolumn{1}{c}{43.66}  & \multicolumn{1}{c}{30.87} \\
    GroundVQA \cite{groundvqa}  &  \multicolumn{1}{c}{26.67}   & \multicolumn{1}{c}{17.63}  & \multicolumn{1}{c}{39.94}  & \multicolumn{1}{c}{27.70} \\    
    EgoVideo \cite{Egovideo2024} &  \multicolumn{1}{c}{25.07}   & \multicolumn{1}{c}{17.31}  & \multicolumn{1}{c}{40.88}  & \multicolumn{1}{c}{29.67} \\
    EgoVideo$^{\dag}$ \cite{Egovideo2024} &  \multicolumn{1}{c}{28.05}   & \multicolumn{1}{c}{19.31}  & \multicolumn{1}{c}{44.16}  & \multicolumn{1}{c}{31.37} \\
    \midrule
    GazeNLQ & \multicolumn{1}{c}{25.24} & \multicolumn{1}{c}{17.58} & \multicolumn{1}{c}{39.99} & \multicolumn{1}{c}{30.24}\\
    GazeNLQ\textsuperscript{$\star$} & \multicolumn{1}{c}{25.45} & \multicolumn{1}{c}{17.48} & \multicolumn{1}{c}{40.23} & \multicolumn{1}{c}{29.87}\\
    \textbf{GazeNLQ}$^{\dag}$ & \multicolumn{1}{c}{27.82} & \multicolumn{1}{c}{18.68} & \multicolumn{1}{c}{43.53} & \multicolumn{1}{c}{30.97}\\
    \bottomrule
    \multicolumn{4}{l}{$^{\dag}$Ensemble results}
  \end{tabular}
  \label{testcomparison}
\end{table}

\begin{table}[t]
  \centering
  \caption{Performance Comparison on NLQ \textit{val} split.}
  \fontsize{8}{9.5}\selectfont
  \begin{tabular}{@{}lc@{}lc@{}lc@{}lc@{}}
    \toprule
    \multicolumn{1}{c}{\multirow{2}{*}{Method}} & \multicolumn{4}{c}{Validation} \\
    \cmidrule(lr){2-5} 
    \multicolumn{1}{c}{} & \multicolumn{1}{c}{R1@0.3} & \multicolumn{1}{c}{R1@0.5} & \multicolumn{1}{c}{R5@0.3} & \multicolumn{1}{c}{R5@0.5} \\
    \midrule
    GroundNLQ \cite{groundnlq2023} & \multicolumn{1}{c}{26.98}   & \multicolumn{1}{c}{18.83}  & \multicolumn{1}{c}{53.56}  & \multicolumn{1}{c}{40.00}  \\
    GroundVQA \cite{groundvqa} &  \multicolumn{1}{c}{29.70}   & \multicolumn{1}{c}{-}  & \multicolumn{1}{c}{-}  & \multicolumn{1}{c}{-}  \\
    EgoVideo \cite{Egovideo2024} & \multicolumn{1}{c}{28.65} & \multicolumn{1}{c}{19.73} & \multicolumn{1}{c}{53.30} & \multicolumn{1}{c}{40.42}  \\
    \midrule
    GazeNLQ & \multicolumn{1}{c}{26.98} & \multicolumn{1}{c}{17.88} & \multicolumn{1}{c}{52.50} & \multicolumn{1}{c}{39.54} \\
    GazeNLQ\textsuperscript{$\star$} & \multicolumn{1}{c}{27.22} & \multicolumn{1}{c}{18.08} & \multicolumn{1}{c}{ 52.61} & \multicolumn{1}{c}{39.63} \\
    \bottomrule
  \end{tabular}
  \label{valcomparison}
\end{table}

\subsection{Performance Comparison}
\cref{testcomparison} reports the comparison results on the NLQ \textit{test} split. Our ensemble approach achieves an R1@0.3 score of 27.82 and an R1@0.5 score of 18.68, demonstrating competitive performance. Notably, the variant incorporating negative gaze embeddings slightly outperforms the standard (positive) gaze formulation. This is an interesting finding that we plan to explore further in future work to understand its implications and potential for enhancing grounding performance.

\cref{valcomparison} presents results on the NLQ \textit{val} split without ensembling. While our method improves the R1@0.3 score compared to GroundNLQ, it results in a slight decrease in the R1@0.5 score. This indicates that our approach is more effective at retrieving relevant segments within a relaxed temporal threshold but less accurate under stricter alignment constraints.

\begin{table}[t]
  \centering
  \caption{Ablation study of whether freeze the weights of Gaze Estimator on NLQ \textit{val} split.}
  \fontsize{8}{9.5}\selectfont
  \begin{tabular}{@{}lc@{}lc@{}lc@{}lc@{}}
    \toprule
    \multicolumn{1}{c}{\multirow{2}{*}{Weights}} & \multicolumn{4}{c}{Validation} \\
    \cmidrule(lr){2-5} 
    \multicolumn{1}{c}{} & \multicolumn{1}{c}{R1@0.3} & \multicolumn{1}{c}{R1@0.5} & \multicolumn{1}{c}{R5@0.3} & \multicolumn{1}{c}{R5@0.5} \\
    \midrule    
    Unfreeze & \multicolumn{1}{c}{26.98} & \multicolumn{1}{c}{18.10} & \multicolumn{1}{c}{51.49} & \multicolumn{1}{c}{38.60} \\
    Freeze  & \multicolumn{1}{c}{27.22} & \multicolumn{1}{c}{18.08} & \multicolumn{1}{c}{ 52.61} & \multicolumn{1}{c}{39.63} \\
    \bottomrule
  \end{tabular}
  \label{tab:ablation}
\end{table}

\subsection{Ablation Study}
We conducted an ablation study to evaluate whether freezing the weights of the pretrained gaze estimation module affects grounding performance in \cref{tab:ablation}. Interestingly, freezing the gaze model's weights results in better performance compared to finetuning. Since the gaze estimator is trained on a relatively small dataset and may not generalize well when finetuned jointly with the grounding model, which is trained on a large-scale narration dataset.

\subsection{Case Analysis}

\begin{figure}[!t]
    \centering

    \begin{subfigure}[b]{\linewidth}
        \centering
        \includegraphics[width=\linewidth]{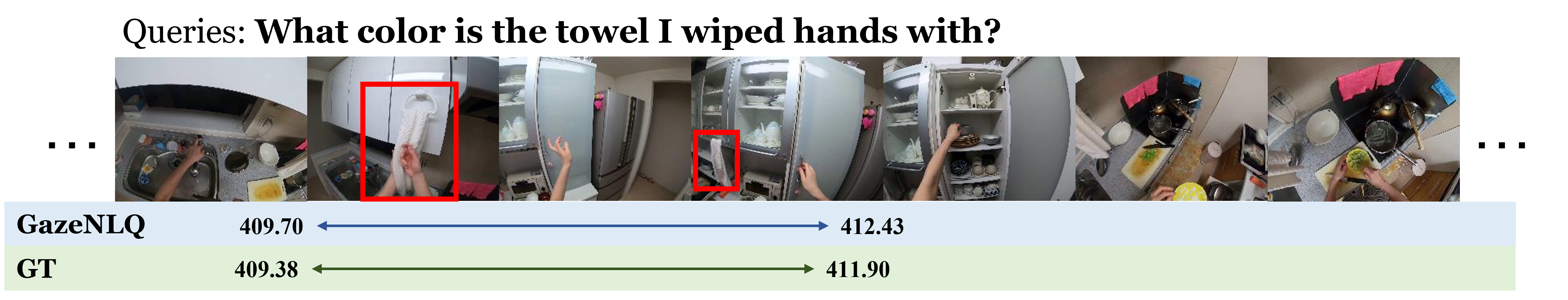}\\[0.5em]
        \includegraphics[width=\linewidth]{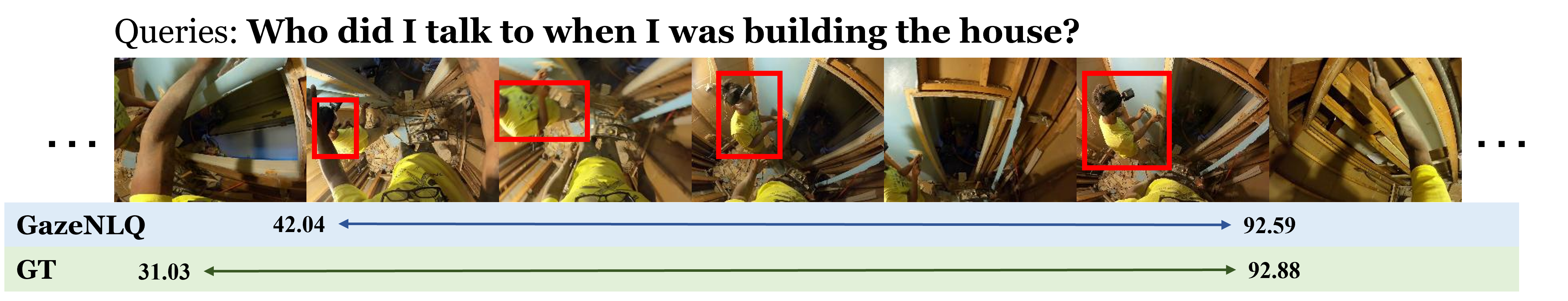}
        \caption{Successful cases}
        \label{fig:success_cases}
    \end{subfigure}
    \vskip 1em
    
    \begin{subfigure}[b]{\linewidth}
        \centering
        \includegraphics[width=\linewidth]{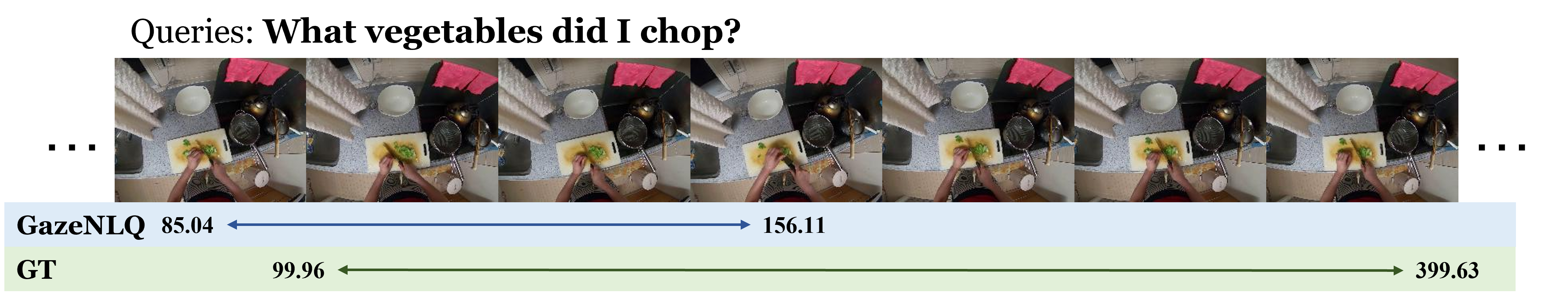}\\[0.5em]
        \includegraphics[width=\linewidth]{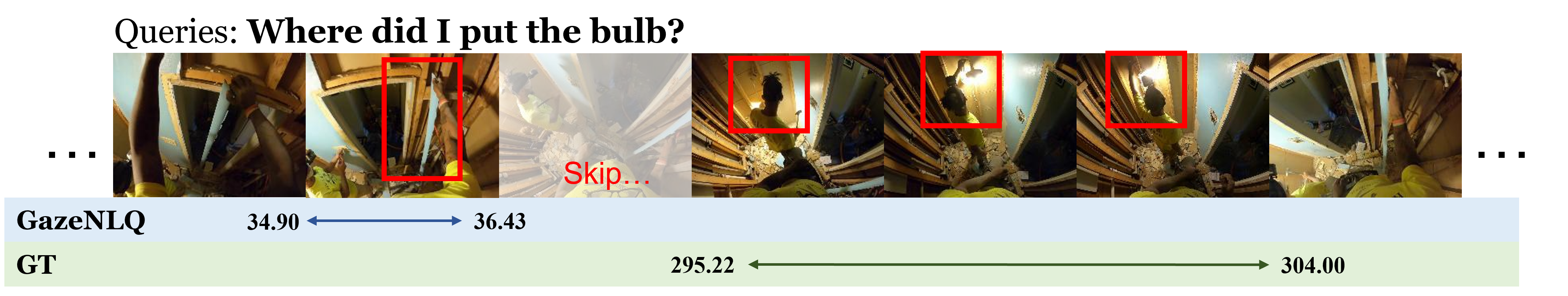}
        \caption{Failure cases}
        \label{fig:failure_cases}
    \end{subfigure}
    \caption{Four examples of GazeNLQ on NLQ \textit{val} split: two successful cases (a) and two failure cases (b).}
    \label{fig:ex_combined}
\end{figure}

\cref{fig:success_cases} shows successful examples in NLQ, where our model accurately locates the target of the text description. However, the failure examples are presented in \cref{fig:failure_cases}. In the top figure, the error arises from an imprecise temporal boundary—GazeNLQ captures only the first half of the ground truth event (“chop the vegetables”), indicating difficulty in handling long-duration actions. In the bottom figure, the model fails due to a misunderstanding of the object involved in the activity. However, we believe the ground truth annotation may not accurately reflect the subject, as the action of placing the bulb was performed by someone other than the camera wearer.

\subsection{Discussion}
This study represents an early stage in our research on egocentric video grounding with gaze, and remains a room for future improvement. First, there exists a feature discrepancy between the gaze training stage and the video grounding stage due to the use of different video feature extractors. The video features for gaze estimator are from Omnivore \cite{omnivore}, while the grounding stage employs \cite{InternVideo2022} video features. This mismatch may hinder the seamless transfer of learned representations, potentially impacting grounding performance.

Second, gaze information serves as a strong spatial prior during the gaze estimation phase, capturing precise locations of visual attention. However, in the grounding stage, the video features lack explicit spatial information. This non-spatial feature structure limits the ability to directly leverage the spatial cues provided by gaze tokens, necessitating additional processing or fusion strategies to align gaze with video features, which may introduce inefficiencies or loss of spatial detail.

Third, our approach relies on finetuning a pretrained GroundNLQ model rather than training from scratch using narration data. This finetuning strategy may constrain the model’s ability to fully adapt to the nuances of our dataset, particularly in integrating gaze information with text queries. Training from scratch with narration data could potentially yield a more robust model but was not pursued due to resource and time constraints at this stage.

\section{Conclusion}
This report presents GazeNLQ, our proposed method for the Ego4D natural language queries challenge at CVPR 2025. GazeNLQ employs a contrastive learning-based pretraining strategy for gaze estimation, which is a core component of the overall framework. The incorporation of estimated gaze into the video representation enhances the model’s ability to localize relevant content in response to natural language queries, as demonstrated by experimental results. These improvements highlight the promise of leveraging gaze to advance egocentric video understanding. Future work will focus on developing consistent feature extractors across stages, incorporating spatial information in grounding features, and exploring training from scratch to enhance model adaptability.
{
    \small
    \bibliographystyle{ieeenat_fullname}
    \bibliography{main}
}


\end{document}